\documentclass{article}
\usepackage{graphicx} 
\usepackage[a4paper, total={4in, 8in}]{geometry}
\usepackage{url}
\usepackage[hidelinks]{hyperref}
\usepackage{listings}
\usepackage[dvipsnames]{xcolor}
\usepackage[
sorting=none
]{biblatex}
\usepackage{listings}

 \title{Transcription and translation of videos using fine-tuned XLSR Wav2Vec2 on custom dataset and mBART \\\line(1,0){\textwidth}}
\begin{document}
\author{
  Aniket Tathe\\
    \texttt{\scriptsize Department of  Mechanical Engineering}\\
  \texttt{\scriptsize MES College of Engineering, Pune, India}\\
   \texttt{\scriptsize \href{mailto:anikettathe.08@gmail.com}{anikettathe.08@gmail.com}}
  \and
  Anand Kamble\thanks{Corresponding author. Email: \scriptsize amk23j@fsu.edu}\\
  \texttt{\scriptsize Department of Scientific Computing}\\
  \texttt{\scriptsize Florida State University, USA}\\
  \texttt{\scriptsize \href{mailto:amk23j@fsu.edu}{amk23j@fsu.edu}}
  \and
  Suyash Kumbharkar\\
    \texttt{\scriptsize Department of Electrical Engineering }\\
    \texttt{\scriptsize and Information Technology}\\
  \texttt{\scriptsize Technische Hochschule Ingolstad, Germany}\\
  \texttt{\scriptsize \href{mailto:suk9387@thi.de}{suk9387@thi.de}}
  \and
  \\
  Atharva Bhandare\\
    \texttt{\scriptsize Department of  Mechanical Engineering}\\
  \texttt{\scriptsize MES College of Engineering, Pune, India}\\
  \texttt{\scriptsize \href{mailto:atharvabhandare512@gmail.com}{atharvabhandare512@gmail.com}}
  \and
  \\
  Anirban C. Mitra\\
    \texttt{\scriptsize Department of  Mechanical Engineering}\\
  \texttt{\scriptsize MES College of Engineering, Pune, India}\\
  \texttt{\scriptsize \href{mailto:amitra@mescoepune.org}{amitra@mescoepune.org}}
}

\date{}

\maketitle
\section{Abstract}
This research addresses the challenge of training an ASR model for personalized voices with minimal data. Utilizing just 14 minutes of custom audio from a YouTube video, we employ Retrieval-Based Voice Conversion (RVC) to create a custom Common Voice 16.0 corpus. Subsequently, a Cross-lingual Self-supervised Representations (XLSR) Wav2Vec2 model is fine-tuned on this dataset. The developed web-based GUI efficiently transcribes and translates input Hindi videos. By integrating XLSR Wav2Vec2 and mBART, the system aligns the translated text with the video timeline, delivering an accessible solution for multilingual video content transcription and translation for personalized voice.

\section{Keywords}
XLSR Wav2Vec2, mBART, RVC, Personalized Speech Conversion, Speaker Diarization.

\section{Introduction}
Training an Automatic Speech Recognition (ASR) model is a formidable task, with the efficacy of the model intricately tied to the quality of the dataset used during training. The challenge becomes even more pronounced when endeavoring to train an ASR model for a low-resource language like Hindi, where the availability of online data, compared to languages such as English, is notably limited. The complexity is further compounded when the aim is to train a personalized ASR model for Hindi, one that not only transcribes but also translates an individual's unique voice. The generation of an appropriate dataset emerges as a significant hurdle in achieving this objective.
This paper confronts the complexities inherent in training personalized ASR models for Hindi by introducing an innovative methodology. Leveraging just 14 minutes of personalized custom audio, a Retrieval-Based Voice Conversion (RVC)\cite{RVC} model is meticulously trained. This RVC model serves as the cornerstone for the creation of a tailored Common Voice 16 corpus, specifically tailored for Hindi. Subsequently, a Cross-lingual Self-supervised Representations (XLSR) Wav2Vec2\cite{babu22_interspeech} model is trained on this bespoke dataset, showcasing adaptability even with limited resources.
To augment the practical utility of the trained model, a user-friendly web-based Graphical User Interface (GUI) is developed. This GUI, powered by Gradio, simplifies the process of transcribing Hindi audio to Hindi text using XLSR Wav2Vec2 and subsequently translating this Hindi text to English using mBART\cite{liu2020multilingual}. The input for this system is a video containing Hindi audio, and the output is a video enriched with accurate English subtitles. This research thus contributes to the advancement of ASR technology for low-resource languages, offering a streamlined pipeline for personalized voice transcription and translation from Hindi audio to English(subtitle), specifically tailored for video content.

\section{Methodology}
\subsection{Common voice 16.0 data augmentation using Ozen toolkit}
The data augmentation is performed on the Common voice 16.0 dataset\cite{ardila2020common}. In the most recent release, Common Voice 16.1, English language data spans approximately 3,438 hours of audio, with 2,586 hours validated by a community of 90,474 contributors and Hindi language section comprises about 21 hours of recorded audio, of which 14 hours have undergone meticulous validation
A 14-minute segment of custom, personalized audio was meticulously extracted from a YouTube video and subsequently converted into the WAV format. Employing the Ozen toolkit\cite{Ozen-toolkit}, a comprehensive processing pipeline ensued, involving speech extraction and transcription using the Whisper module. The transcribed results were then saved in the LJ format. It is noteworthy that the audio files are systematically stored in the designated 'wavs' folder. Simultaneously, the transcribed Hindi texts are organized into separate 'train' and 'valid' text files, facilitating effective dataset management.
It is crucial to emphasize that Whisper primarily defaults to English transcription. To adapt this framework for Hindi transcription, necessary modifications were made within the utils.py file, specifically altering the "task" to "transcribe" and setting the "language" to "hi". These adjustments ensure accurate and contextually relevant transcription for the Hindi language.
Additionally, the Ozen toolkit incorporates the pyannote\cite{Bredin2021} framework for speaker diarization, further enhancing its utility in speech and audio processing applications. This integration enables precise identification and separation of speakers within the audio, contributing to the overall robustness of the system. 

\subsection{RVC model training and inference}
\begin{figure}[ht]
    \centering
    \includegraphics[width=0.5\textwidth]{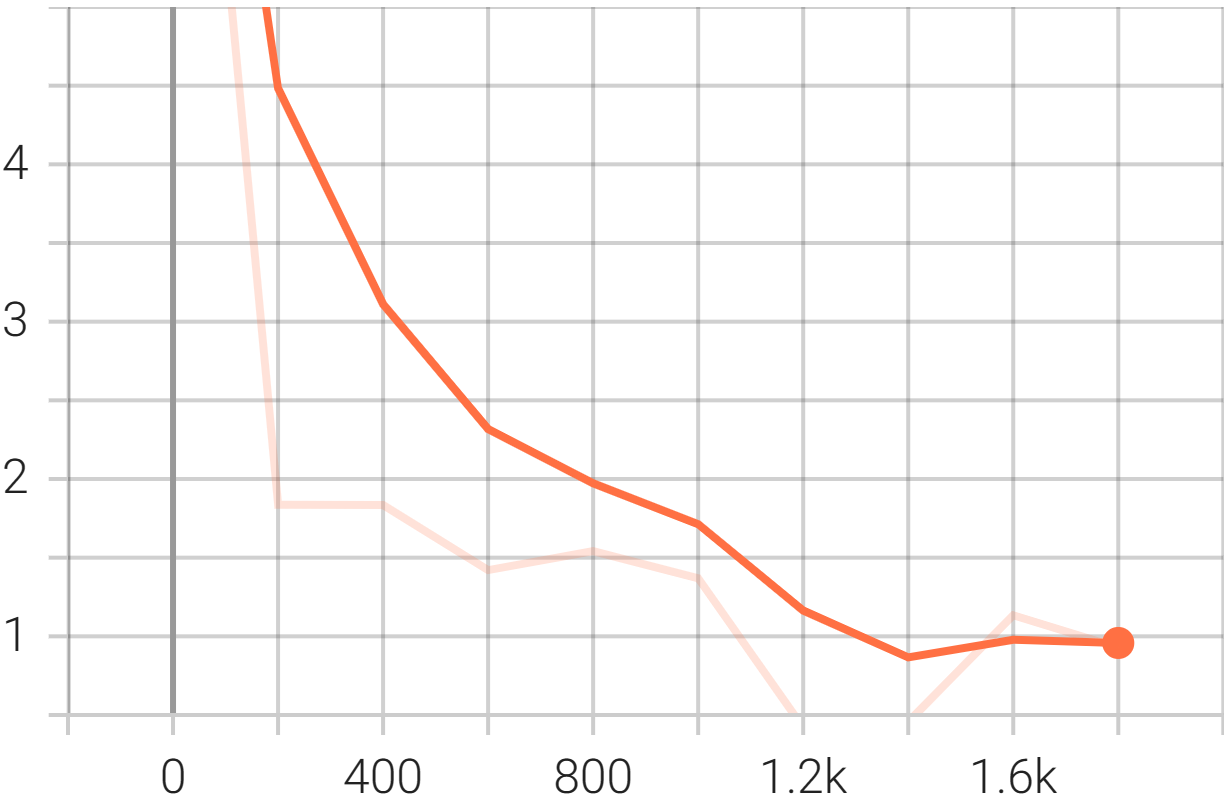} 
    \caption{KL Divergence Loss (loss\_kl)}
    \label{fig:1}
\end{figure}

The Retrieval-based-Voice-Conversion-WebUI by RVC-Project is an open-source tool enabling users to convert voice data using a retrieval-based voice conversion approach. It is freely accessible on GitHub [14] under the MIT software license. Recommended for audio recordings of a minimum 10-minute length for optimal model training, the project offers noise removal capabilities through Vocals/Accompaniment Separation and Reverbation, utilizing the 'HP2-all-vocals' model. The target sample rate is set at 32,000, and the training utilizes the base v2 pre-trained model (f0G32K and f032K) with a batch size of 40 for 200 epochs on the NVIDIA A5000 GPU. Post 200 epochs, the KL Divergence Loss reaches 0.9292. This can be viewed in \hyperref[fig:1]{Fig.1}  KL Divergence Loss (loss\_kl) below.

\begin{figure}[ht]
    \centering
    \includegraphics[width=0.7\textwidth]{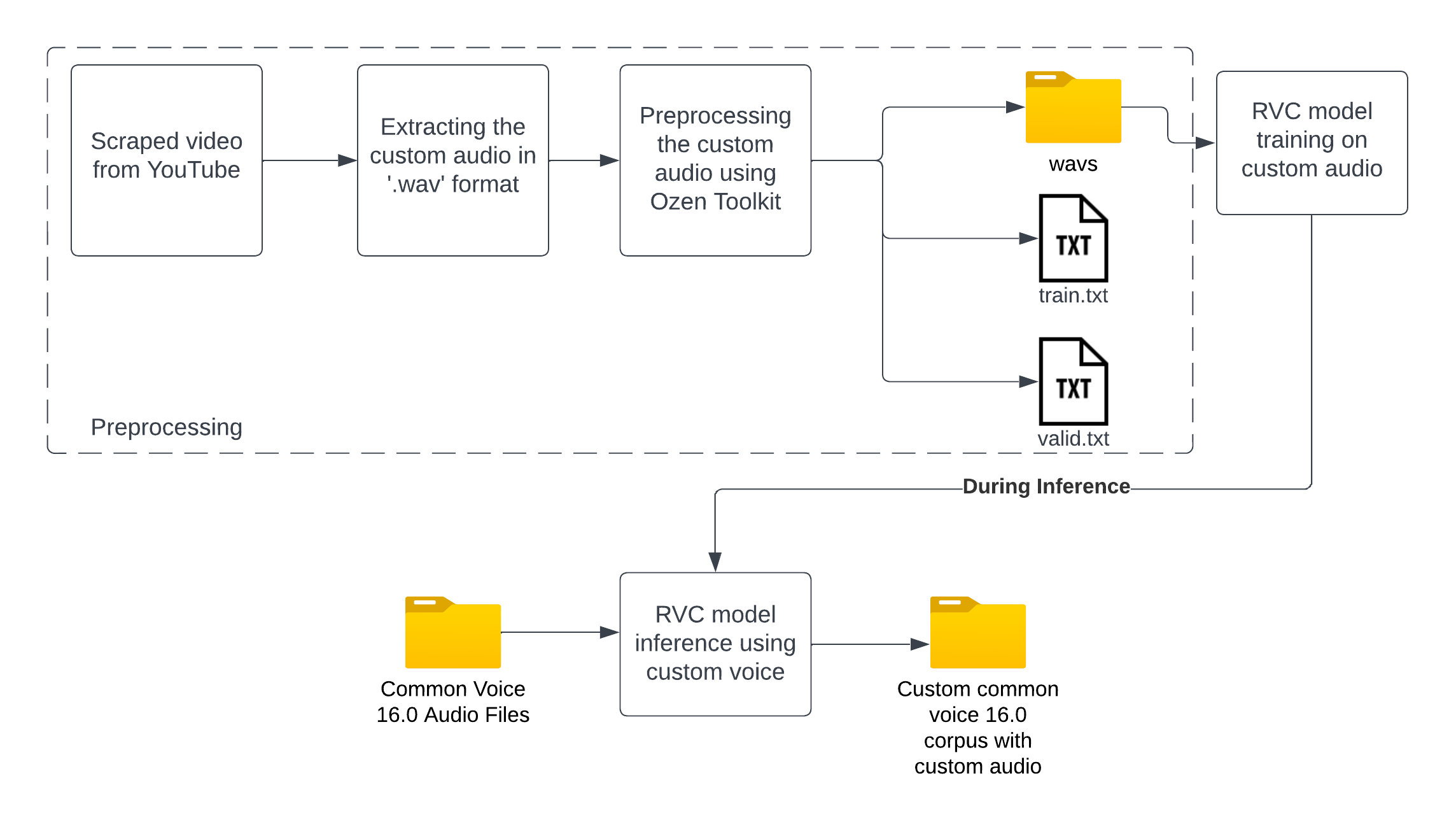} 
    \caption{Data Augmentation}
    \label{fig:2}
\end{figure}

After completing the training phase, the model underwent inference to reproduce the acquired voice characteristics. Utilizing Common Voice audio inputs alongside the trained model's inferencing, coupled with specified index paths, a bespoke common voice dataset was generated, showcasing the distinctive custom voice. Critical parameters such as volume envelope scaling (0.25), filter radius (3), and search feature ratios (0.75) and 0.33 for safeguarding voiceless consonants and breath sounds were implemented. It is imperative to note that these parameter values may necessitate adjustment based on the audio utilized during training, and users are encouraged to experiment with different settings to identify the optimal configuration. The resulting customized Common Voice 16.0 dataset, featuring the unique custom voice, is accessible under 
" Aniket-Tathe-08/Custom\_Common\_Voice\_16.0\_dataset\_using\_RVC\_14min\_data" on Hugging Face. The pipeline can be viewed in \hyperref[fig:2]{Fig.2} Data Augmentation below.

At the end of this process, we have generated a custom common voice 16.0 dataset with input audio which only comprises the personalized voice with the help of RVC which will be further used to fine tune the XLSR model.

\subsection{Fine-tuning XLSR Wav2Vec2 model}
XLSR-Wav2Vec2 is a framework for self-supervised learning of speech representations proposed by Facebook AI Research. It builds on the original Wav2Vec \cite{schneider2019wav2vec} but is trained on much larger speech recognition datasets to learn more robust representations. Specifically, XLSR-Wav2Vec2 leverages the self-supervised objective of masked reconstruction to pre-train a deep convolutional neural network model that encodes speech audio inputs into latent speech embeddings. More than 56,000 hours of unlabeled speech data across 50 languages are utilized to pre-train the XLSR model. This allows it to learn universal speech representations capturing various linguistic properties. Fine-tuning the pre-trained model on downstream speech tasks requires only a few minutes of labeled data to achieve state-of-the-art performance on benchmarks ranging from speech recognition to speech translation. The encoder model architecture and self-supervised pre-training technique allow XLSR-Wav2Vec2 representations to generalize very effectively across languages and domains.

\begin{figure}[ht]
    \centering
    \begin{tabular}{cc}
    \includegraphics[width=0.4\linewidth]{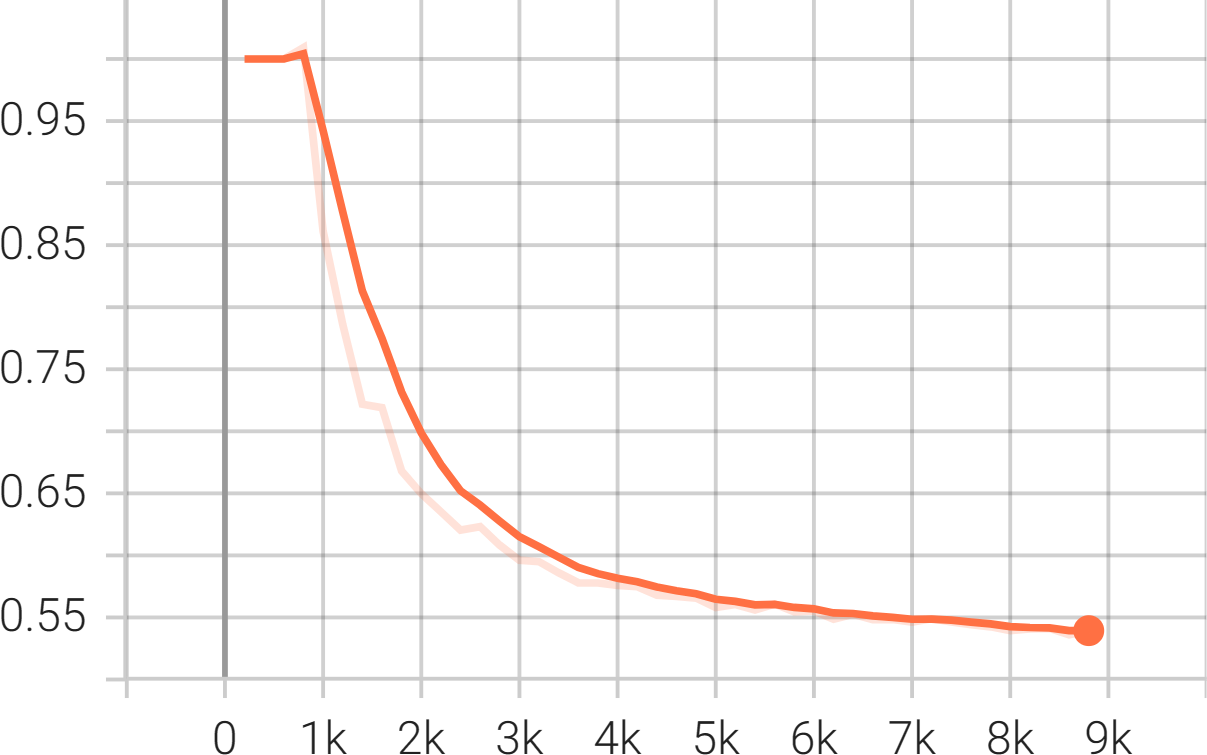} & 
    \includegraphics[width=0.4\linewidth]{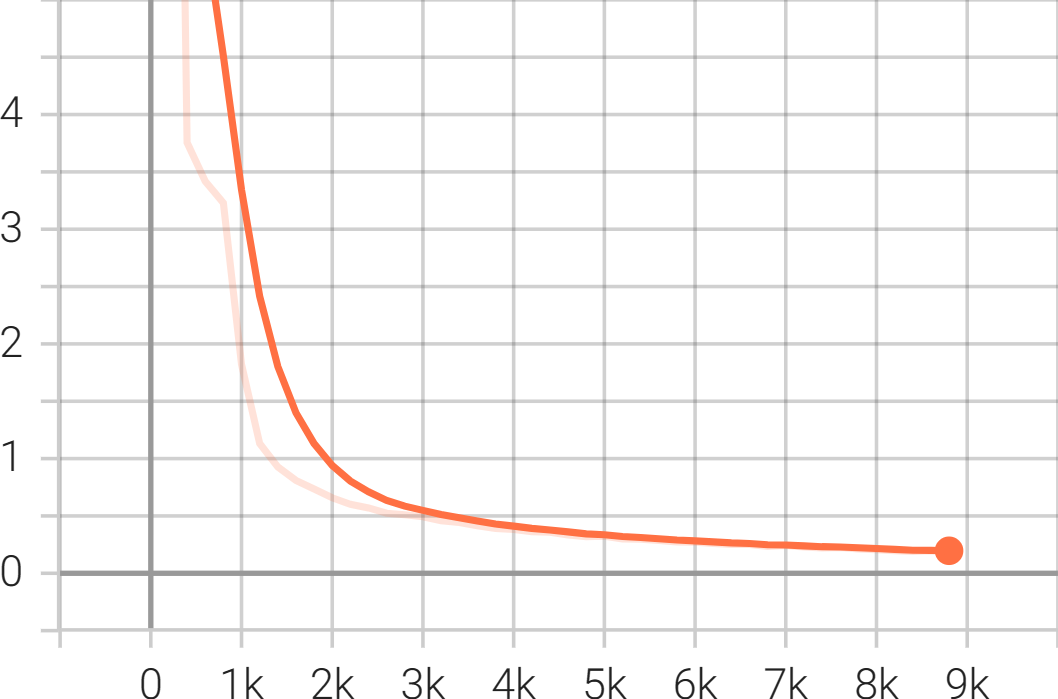}  \\
    (a) WER & (b) Training Loss
    \end{tabular}
    \label{fig:6}
\end{figure}

”facebook/wav2vec2-large-xlsr-53” was the model used for fine-tuning and the training process occurred on an NVIDIA A5000 GPU for 40 epochs with a learning rate of $1 \times 10^{-4}$ and a weight decay of $2.5 \times 10^{-6}$ Although there wasn't a significant difference, the model exhibited slightly better performance with weight decay. The model gave a training accuracy of 0.80 and WER of 0.53 The WER graph and training loss can be seen below in \hyperref[fig:6]{Fig.3}. It was seen that the model was overfitting on the training data. This model trained on a personalized custom Common Voice 16.0 dataset can be found at  ” Aniket-Tathe-08/XLSR-Wav2Vec2-Finetuned-14min-dataset”, which accepts Hindi audio input and generates corresponding Hindi text. \\
\vspace{20pt}

\subsection{Neural machine translation using mBART}
mBART (multilingual Bidirectional and Auto-Regressive Transformer) (Liu et al., 2020) is a multilingual sequence-to-sequence model pre-trained using a denoising autoencoding approach on large-scale monolingual corpora spanning 25 languages. It employs a standard Transformer-based encoder-decoder architecture (Vaswani et al. 2017) and is pre-trained to reconstruct input text fragments that have been corrupted through an arbitrary noising function. This self-supervised pre-training enables mBART to learn universal linguistic representations that transfer across multiple languages and downstream tasks. After pre-training on around 200GB of text data, mBART can be fine-tuned on target sequence generation tasks by simply adding task-specific decoder heads. Requiring only a few thousand labelled examples in a low-resource setting, fined-tuned mBART has achieved state-of-the-art performance on various multilingual benchmarks including translation, summarization, and question answering. The pre-trained representations and adapted parameters effectively encapsulate multilingual sequences, allowing mBART to generalize well even for low-resource language generation tasks.

\subsection{Speaker Diarization using pyannote/speaker-diarization-3.1}
Pyannote is an open-source speech processing toolkit designed for constructing speaker diarization systems. Diarization addresses the question of 'who spoke when?' in an audio recording. In the end-to-end pipeline, Pyannote determines the start and end times of each speaker, leading to the segmentation of the input audio into multiple segments with the format (start, end, speaker). This information is then used to sequentially process the cropped audio segments through XLSR Wav2Vec2 and mBART for subtitle generation.
The output is stored in an 'output.vtt' file, containing both the start and end times based on the diarization and the output generated by XLSR Wav2Vec2 and mBART. An example of the 'output.vtt' file is illustrated in the \hyperref[lst:1]{Listing 1}  Diarization output below.

\begin{lstlisting}[caption={Diarization output},label={lst:1}]
00:00.000 --> 00:06.400
So now let's come to Japan. When I came to Japan, I was about 22 years old and I have been living here for 8 years.

00:06.400 --> 00:10.400
And here some people are curious about the salary, how much salary is there in Japan.

00:10.400 --> 00:32.400
So personally, I feel that according to my salary, there is no problem in social media.
\end{lstlisting}

\subsection{Web GUI using Gradio}

\begin{figure}[ht]
    \centering
    \includegraphics[width=0.9\textwidth]{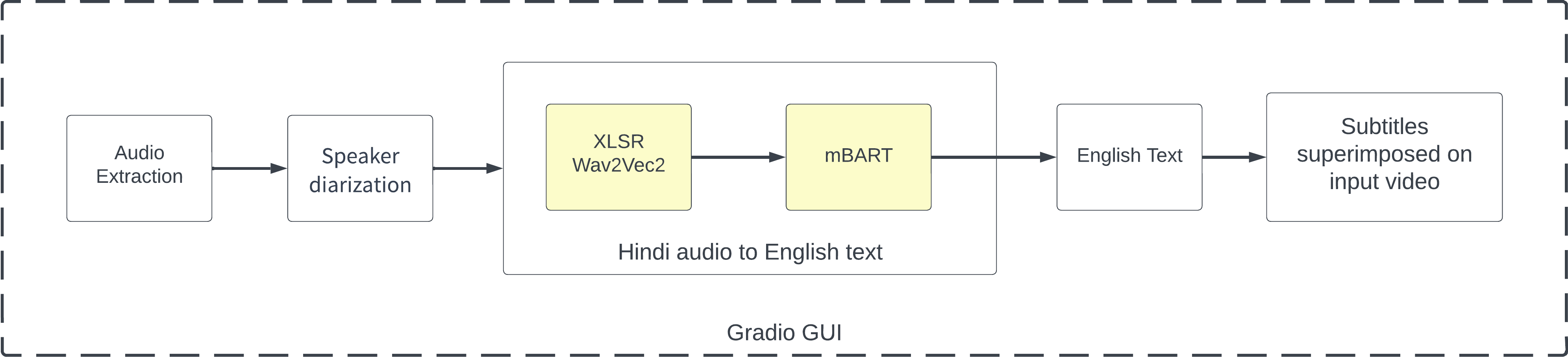} 
    \label{fig:4}
\end{figure}

After uploading the input video with Hindi audio to the GUI, an 'mp3' file is generated, containing the extracted Hindi audio from the video. This 'mp3' file is then input into Pyannote for speaker diarization. The output of Pyannote, in the format (start, end, speaker), is used to crop the audio accordingly using this ‘start’ and ‘end’. The cropped audio segments are then processed through the XLSR Wav2Vec2 model for Hindi audio to Hindi text conversion and mBART for translating Hindi text to English. Subsequently, subtitles are generated. Using the 'start', 'end', and subtitle results, an 'output.vtt' file is created. This 'output.vtt' file is then overlaid onto the input video, producing the final output video with subtitles, as illustrated in the pipeline below in \hyperref[fig:4]{Fig. 4}.
\vspace{20pt}

\begin{figure}[ht]
    \centering
    \includegraphics[width=1.0\textwidth]{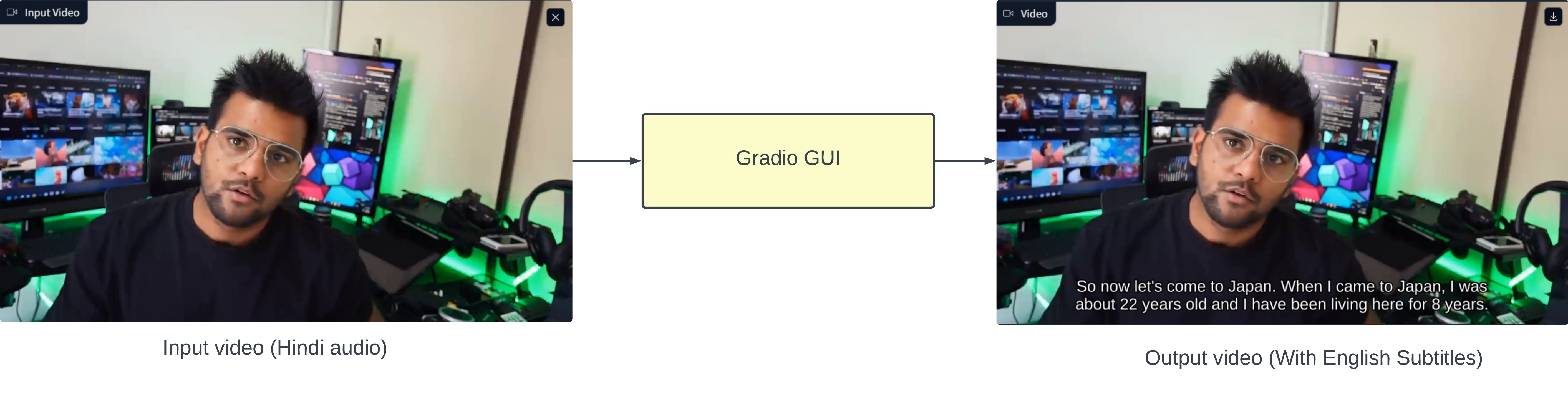} \\
    Figure 4: End-to-end pipeline
    \label{fig:5}
\end{figure}

\newpage
\section{Results and Discussion}
Training a custom ASR model for personalized audio in a low-resource language like Hindi remains a significant challenge, requiring further research in this domain. The XLSR Wav2Vec2 model can be fine-tuned to achieve better accuracy, enabling more precise transcriptions for personalized audio. Using just 14 minutes of personalized custom audio, an RVC model was trained to generate a custom Common Voice 16.0 corpus. 
This corpus was then employed to train an XLSR Wav2Vec2 model, resulting in approximately 0.80 accuracy and 0.53 WER. Alternatively, one can explore using other datasets like LibriSpeech\cite{7178964}. This method not only facilitates data augmentation but also enables the training of a personalized custom ASR model with high accuracy, even with a very limited dataset. \\
\vspace{10pt}
\hrule

\nocite{9906230}
\nocite{yi2021applying}
\nocite{10095644}
\nocite{shahgir2022applying}
\nocite{peng21e_interspeech}
\nocite{10095036}
\nocite{smith2018disciplined}

\printbibliography[title={References}]
\end{document}